%% file: colm2026_conference.tex
\documentclass{article} 
\usepackage[preprint]{colm2026_conference}
\input{math_commands.tex}

\usepackage{microtype}
\usepackage{hyperref}
\usepackage{url}
\usepackage{booktabs}

\usepackage{subcaption}
\usepackage{hyperref}
\usepackage{url}
\usepackage{booktabs}
\usepackage{multirow}
\usepackage{xcolor}
\usepackage{rotating}
\usepackage[table]{xcolor}

\definecolor{verylightgray}{gray}{0.9}
\newcommand{\gcell}{\cellcolor{verylightgray}}
\definecolor{darkgreen}{rgb}{0,0.5,0}
{}

\newcommand{\best}[1]{\textcolor{darkgreen}{\underline{\textbf{#1}}}}

\newcommand{\secondbest}[1]{\underline{\textbf{#1}}}

\usepackage{lineno}

\definecolor{darkblue}{rgb}{0, 0, 0.5}
\hypersetup{colorlinks=true, citecolor=darkblue, linkcolor=darkblue, urlcolor=darkblue}

\title{Code Comprehension then Auditing for Unsupervised LLM Evaluation}

\author{Bhrij Patel \\
Department of Computer Science\\
University of Maryland, College Park\\
\texttt{bbp13@umd.edu} \\
\And
Souradip Chakraborty \\
Department of Computer Science\\
University of Maryland, College Park\\
\And
Mengdi Wang \\
Department of Computer Science \\
Princeton University 
\And
Dinesh Manocha \\
Computer Science Department \\
University of Maryland, College Park  \\
\AND
Amrit Singh Bedi \\
Computer Science Department \\
University of Central Florida
}

\begin{document}

\ifcolmsubmission
\linenumbers
\fi

\maketitle

\begin{abstract}
Large Language Models (LLMs) for unsupervised code correctness evaluation have recently gained attention because they can judge if code runs as intended without requiring reference implementations or unit tests, which may be unavailable, sparse, or unreliable. However, most prior approaches condition LLM evaluators directly on the full code implementation, forcing the model to jointly infer program behavior and evaluate correctness in a single step. This entanglement leads to misinterpretations of code behavior and unreliable judgments. To mitigate this issue, we introduce \textbf{CoCoA}, an unsupervised \textbf{Co}de \textbf{Co}mprehension then \textbf{A}uditing framework that first comprehends functionality to generate a natural-language explanation. Then it evaluates task alignment based on this explanation. By sequentially sampling comprehension before evaluation, CoCoA improves the quality of inferred program behavior and enables the evaluator to focus on behavioral alignment rather than raw implementation details. Across multiple datasets, programming languages, and models, CoCoA achieves up to 68\% increased F1 score and up to 20\% increased accuracy over the best-performing baselines. 
\end{abstract}

\section{Introduction}

Unsupervised evaluation of code with Large Language Models (LLMs) \citep{achiam2023gpt, yang2025qwen3} has recently gained attention due to its scalability and practicality by not requiring reference solutions or unit tests \citep{tong2024codejudge, zhuo2024ice, ugare2026agentic, wang2025mctsjudgetesttimescalingllmasajudge, jin2025uncovering}. In this setting, a frozen LLM during test-time must determine whether a program satisfies a natural-language specification solely from its implementation. This evaluation requires reasoning about the program’s functional behavior. Without any supervised signal, an evaluation framework must rely on its own internal understanding of the code behavior to evaluate task alignment \citep{wang2025mctsjudgetesttimescalingllmasajudge, jin2025uncovering}. Prior work has highlighted the issue of false negatives (incorrectly flagging code as ``buggy'') in evaluation \citep{tong2024codejudge, jin2025uncovering}. Both false positives and false negatives incur costs in real-world software development, making metrics such as F1 important to optimize.

Most prior unsupervised evaluation methods directly condition LLM judges on the raw code implementation \citep{tong2024codejudge, zhuo2024ice, jin2025uncovering}. Thus, comprehension and evaluation are implicitly entangled within a single LLM sampling step. Recent studies show that such approaches often fail to correctly reason about program logic \cite{tong2024codejudge} and are sensitive to misleading comments \citep{moon2025don, lamcodecrash}. Such issues reduce the reliability of LLM-based evaluation pipelines and can negatively impact workflows such as code refinement \citep{dong2024self, yuksekgonul2025optimizing}. 

To address this limitation, we draw inspiration from the forward process of code development. When humans or AI agents solve programming tasks, they typically first form a high-level plan, or pseudocode \citep{yu2025recode}, describing the program logic before implementing the code. In practice, such as in manual software development, coding interviews, and AI-assisted development tools \citep{antigravity2025}, these plans are often evaluated for correctness before the final implementation is written. Motivated by this observation, we hypothesize that the high-level plan is a sufficient representation for determining code alignment with a task. However, during evaluation, this plan is unavailable. 

Motivated by this idea, we treat the program’s high-level logic as a latent variable that must be inferred from code. Consequently, we introduce \textbf{CoCoA:} \textbf{C}ode \textbf{C}omprehension then \textbf{A}uditing. CoCoA first employs an explainer LLM to abstract the implementation into a natural language description of its program behavior. The LLM judge then evaluates alignment based on this explanation, rather than the raw code. By decoupling the comprehension from evaluation, CoCoA provides a more accurate understanding of functionality and thus a more reliable judgment of correctness. Specifically, by sequentially sampling the explanation from an LLM, we allow it to focus on understanding high-level behavior from low-level implementation details. Then, evaluating on this explanation space allows the LLM judge to focus on whether the inferred behavior aligns with the task requirements.  Furthermore, we have the added benefit of abstracting away syntactic variations, leading to a robustness to misleading comments shown in prior work \citep{moon2025don, lamcodecrash}. 

Unlike other test-time methods \citep{tong2024codejudge, jin2025uncovering}, where intermediate reasoning is used only in the latent space within a single call, our approach explicitly replaces the evaluation input with a generated explanation and performs judgment conditioned only on this representation. To the best of our knowledge, this is the first work to explicitly propose evaluating code alignment through an inferred intermediate plan representation. Our main contributions are as follows:

\begin{itemize}
    \item \textbf{Simple, Modular Solution:} We introduce CoCoA, a novel method that generates an intermediate natural language explanation that sufficiently represents code functionality and logic. To the best of our knowledge, we are the first work to propose evaluating on this explanation space, rather than the implementation space, to decouple comprehension from evaluation.

    \item \textbf{Reliable Code Correctness Evaluation and Robustness:} Across $9$ datasets, programming languages, and models, we show that CoCoA has higher accuracy and F1 score on correctness evaluation over baselines that directly judge based on implementation. We also show robustness to misleading comments that prior literature has revealed to be a weakness of LLM judges. 

    \item \textbf{Analysis and Downstream Application:} We provide in-depth experimentation that ablates on key components of CoCoA and the improved benefits of CoCoA-based feedback over baselines for code refinement evaluations.

\end{itemize}

\begin{figure}
    \centering
    \includegraphics[width=\linewidth]{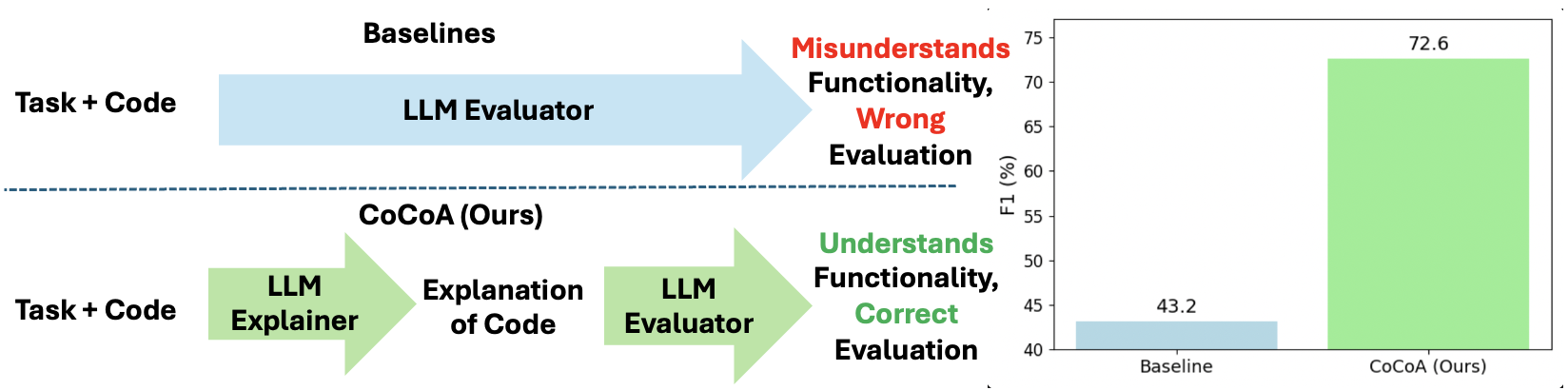}
    \caption{\textbf{Code Comprehension then Auditing.} Unsupervised LLM judges often directly evaluate code by reasoning over implementation details. We propose that first explaining the code to focus on functionality comprehension and auditing based on the explanation leads to more reliable and accurate evaluation.}
    \label{fig:intro}
\end{figure}

\section{Background}

Test-based methods such as pass@k \citep{kulal2019spoc, chen2021evaluating} use unit test execution results to evaluate given code snippets. Other work has developed token-based methods used for machine translation and applied them to evaluate code against a reference \citep{ren2020codebleu, ruby}. Furthermore, embedding-based methods compare the embeddings of the candidate code snippet to a reference code  \citep{zhou-etal-2023-codebertscore}. 
LLM-as-a-Judge \citep{zheng2023judgingllmasajudgemtbenchchatbot, li2024generation}, has been growing in interest due to the ability of LLMs to evaluate large outputs like text \citep{sellam-etal-2020-bleurt, kocmi-federmann-2023-large} quickly and to align with human preferences. \citet{zhuo2024ice} introduced using LLM-generated feedback for code evaluation, and \citet{tong2024codejudge} investigated how LLM judges without references or execution results can evaluate code by first generating an evaluative analysis of the code and then compressing it into a binary ``Correct/Incorrect'' answer.  \citet{jin2025uncovering} also prompted the LLM to analyze and evaluate the code in a single call, and they specifically focused on how to reduce the false negative rate that was previously seen in \citet{tong2024codejudge}. In this work, we show that an LLM-based judge for code correctness becomes unreliable when needing to directly evaluate on the implementation, thus motivating CoCoA to summarize the code implementation in natural language. We also look to establish a \textit{balanced} classifier and analyze the effects on false positives and false negatives. Furthermore, LLMs for understanding and summarizing code has been useful for applications such as creating documentation and comments \citep{nam2024using, sun2025source, khatib2026examiningllmsabilitysummarize} or evaluating code changes \citep{lin2025codereviewqa}.  Recently, \citet{shi2026codeocr} investigated code understanding from images. In this work, we claim that a code understanding module should take place before downstream correctness evaluation. To the best of our knowledge, we are the first work to apply a separate code comprehension module to evaluation.



\subsection{Problem Formulation}

\paragraph{Prompt Optimization for Correctness Evaluation.} Let $t$ be a natural language coding task and let $c$ be a candidate code snippet to solve $t$. Let $a_{c,t} \in [0, 1]$ be the binary ground-truth correctness label. Let $\pi_j$ be a probability distribution parameterized by a frozen LLM that predicts the alignment of $c$ for $t$. Specifically, $ \pi_j(\cdot | p)$ is a conditional distribution based on input prompt $p$. We thus can write our evaluation problem with the prompt optimization objective,

\begin{equation}\label{eq:obj}
    \max_p \sum_{c,t} \mathbb E_{\hat a \sim \pi_j(\cdot | p)}[\mathbf1_{a_{c,t} = \hat a}].
\end{equation}

While Equation \ref{eq:obj} corresponds to maximizing accuracy, it does not distinguish between different judgment error types. In particular, a classifier that trivially predicts all programs as incorrect or all as correct can achieve reasonable accuracy under imbalanced distributions despite being uninformative. Therefore, beyond accuracy, we also seek balanced evaluators that can reliably identify correct and incorrect code. 

\begin{figure}
    \centering
    \includegraphics[width=\linewidth]{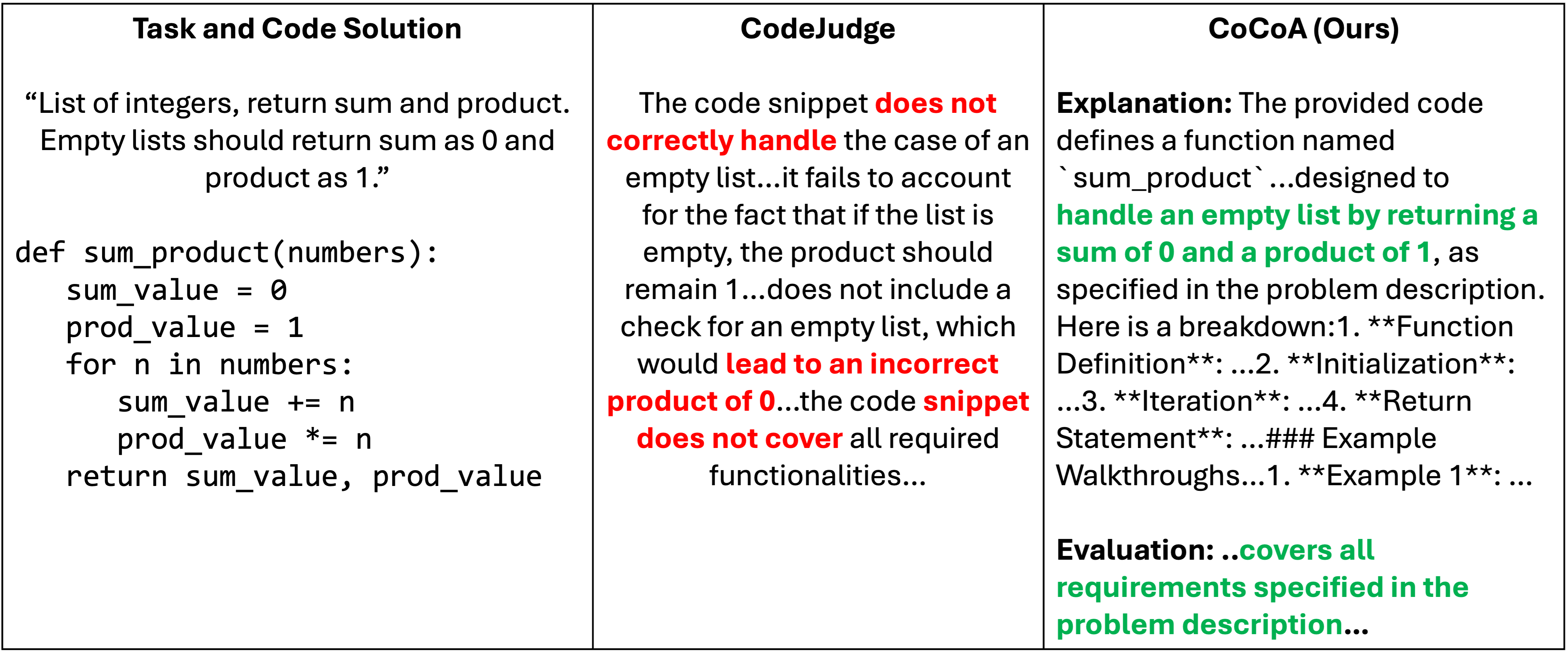}
    \caption{\textbf{Representative Case: Better Code Comprehension Leads to Better Evaluation.} In this example, baseline methods (\textcolor{red}{red}) incorrectly penalize the solution for misunderstanding the logic of the program. CoCoA (\textcolor{darkgreen}{green}) first explains the program to recover a natural-language description of its logic, and then evaluates whether this logic satisfies the task requirements. By judging the explanation rather than the raw code, CoCoA recognizes that the empty-list case is handled by the implementation.}
    \label{fig:example_case}
\end{figure}

\paragraph{Limitations of Prior Work: Evaluation Prompting with Implementation.} Typically, the input evaluation prompt is comprised of the task and the code implementation, $p = (t, c)$. On the left of Figure \ref{fig:example_case}, the canonical solution to the code takes initializes the sum and product variables to $0$ and $1$, respectively, thus handling the empty list case without an explicit check. Evaluation methods with $p = (t,c)$ implicitly ask the LLM evaluator to jointly comprehend and evaluate the code for correctness. Let $e$ be a natural language understanding of the code. Then, we can formally write $(e, \hat a) \sim \pi_j(\cdot|t,c)$. This joint sampling tends to lead to unreliable understanding and thus unreliable evaluations. In the middle of \ref{fig:example_case}, CodeJudge from \citet{tong2024codejudge} misunderstands and thinks the product value will be incorrect with an empty list input.

\section{Proposed Approach: Decoupling Understanding from Evaluation}
\paragraph{Key Idea: Code Explanation is a more Explicit Representation of Task Alignment.} We hypothesize that alignment between code and a task is determined through a high-level explanation of the program logic. This reflects the forward process of software development, where implementations are produced by translating a conceptual plan into executable code. While the implementation $c$ contains many low-level details, the aspects relevant to behavioral correctness are captured by a higher-level logic plan $z$. Formally, $c = I(z)$ where $I$ is an implementation function, and $c$ is a realization of $z$. Thus, the alignment $a_{c,t}$ between $c$ and $t$ can be evaluated through $z$ and $t$. If we had access to an oracle plan $z$ for $c$, we could then construct a prompt $p = (t, z)$. Thus, $\pi_j$ would not need to understand both functionality from implementation and alignment from high-level logical flow. However, $z$ is unavailable in our unsupervised setting. Therefore, we look to generate an \textit{explicit} explanation $e$ of the code implementation. We hypothesize that a faithful explanation is a better representation for correctness judgment than raw implementation, as it better explicitly captures behavior.
\paragraph{CoCoA: Code Comprehension then Auditing.}  
We now present CoCoA as a simple and practical instance of our explanation then evaluation paradigm. We sample an \textit{ explicit} explanation $e$ of implemented code $c$. Let $\pi_e$ be the explanation agent that samples a natural language explanation $e$ from a conditional distribution given a code snippet $c$ for task $t$. We prompt the LLM parameterizing $\pi_e$ to understand the code logic and output an objective explanation of the $c$. Then let $\pi_j$ be the evaluation agent that samples a binary alignment evaluation $\hat a$ from a distribution conditioned on prompt $p = (t, e)$. Formally, $e \sim \pi_e(\cdot|t, c), \;\hat a \sim \pi_j(\cdot|t, e)$. The bottom of Figure \ref{fig:intro} illustrates this paradigm. By first generating an explanation without worrying about evaluation, the LLM can focus solely on understanding the code functionality and hypothetical input/output cases. The right of Figure \ref{fig:example_case} demonstrates our CoCoA approach. CoCoA correctly identifies that the code handles empty lists and provides a code breakdown with an example walkthrough. The evaluator then correctly audits the code based on this $e$.

In this work, we have five different prompts that we experiment with to instruct $\pi_e$. Each one specifies a different level $\lambda \in [1, 5]$ of implementation details to incorporate into the explanation. $\lambda = 1$ prompts the explainer to give a general overview of the code, and $\lambda = 5$ instructs the explainer to be detailed in its explanation and consider unhandled edge cases. We provide prompts in Appendix \ref{sec:prompts}.





\section{Experimental Results}

We conduct experiments to answer: \textbf{(RQ1, Main Results)} How does decoupling understanding and evaluation impact code judgment reliability? \textbf{(RQ2, Ablations)} What is the impact of the different steps in CoCoA? \textbf{(RQ3)} How does CoCoA-based feedback affect downstream code refinement?

\subsection{Setup}




\paragraph{Data Collection and Generation.} Per coding task $t$, we test our evaluation baseline and methods on the four code implementations: 1) the canonical solution $c_t^*$; 2) an incorrect solution $c_t^-$; 3) $c_t^*$ with misleading comments; and 4) $c_t^-$ with misleading comments. We have $9$ datasets: \textbf{ 1-5) HumanEval-X (HE)} \citep{zheng2023codegeex}: a collection of datasets that have the same coding tasks and a solution in different languages, Python (HE-PY), C++ (HE-CPP), Go (HE-Go), Javascript (HE-JS), Java (HE-JA); \textbf{6) APPS} \citep{hendrycksapps2021}; \textbf{7) BigCodeBench (BCB)} \citep{zhuo2024ice}; \textbf{8) DebugBench (DB)} \citep{tian-etal-2024-debugbench};  \textbf{9) VibePass (VP)} \citep{bansal2026vibepass}. Each dataset contains a solution $c_t^*$.\footnote{VibePass only contains ``silver solutions'' that pass all test cases but are not guaranteed to be absolutely correct.} Only DB and VP contain pre-provided ``buggy implementations'' that we treat as $c_t^-$. For the HE\footnote{We provide our version of HE-PY in the supplementary}, APPS, and BCB datasets, we perturb each canonical solution with an LLM (\textit{GPT-4o-mini}) to make subtle changes to $c_t^*$ to obtain the functionally incorrect snippet $c_t^-$. We strip comments from both $c_t^*$ and $c_t^-$ to ensure no hints can be given for code functionality. For all datasets, we generate versions of both $c_t^*$ and $c_t^-$ with \textit{misleading task bias comments} that incorrectly describe the functionality of different portions of the code using the prompt from \cite{moon2025don}.  We chose this type of perturbation as it directly interferes with understanding the code functionality. We provide more details on the datasets and the number of tasks used per dataset in Appendix \ref{sec:app_exp_details}. 

\vspace{-3mm}

\paragraph{Baselines.} All baseline methods evaluate correctness with implementation $c$ as a direct input. We compare against a \textbf{Vanilla} evaluation call that takes in the entire snippet $c$ and is prompted to give a binary assessment on correctness with reasoning. We also compare with unsupervised methods from prior literature: \textbf{ICE-Score} \citep{zhuo2024ice}, \textbf{CodeJudge} \citep{tong2024codejudge}, \textbf{MCTS-Judge}\footnote{due to the amount of sequential calls of MCTS-Judge and computational resources, we only compare with MCTS-Judge using GPT-4o-mini on a subset of the datasets.} \citep{wang2025mctsjudgetesttimescalingllmasajudge}, \textbf{Two-Phase Reflection (2PR)} \citep{jin2025uncovering}, and \textbf{Behavior Comparison (BC)} \citep{jin2025uncovering}. This set of baselines is a mix of single-call and multi-call methods. To more clearly highlight the improvements achieved by separating comprehension and evaluation into separate, sequential calls, we also create another baseline, \textbf{Comprehension then Auditing-Single Prompt (CA-SP)}, that modifies the BC prompt to comprehend code, then evaluate within a single LLM call. Further details on these baselines are provided in the Appendix. 

\vspace{-3mm}

\paragraph{Metrics.} For code evaluation, we report the \textbf{Accuracy (Acc)} and \textbf{F1} score of each method.  We discuss metrics of our code refinement experiments in Section \ref{sec:cr}.

\vspace{-3mm}

\paragraph{LLM Backbones.} We use \textit{GPT-4o-mini} to perturb the canonical solutions for the incorrect solutions and the adversarial comments, and we use it as the code refiner agent. For each evaluation method, we use the same LLM backbone across all steps. We experiment with closed-source (\textit{GPT-4o-mini}) and open-source (\textit{Qwen-2.5-7B-Turbo} and \textit{Llama-3.1-8B-Instruct-Turbo}) models for the evaluation methods. 


\begin{table}[!h]
\centering
\small
\setlength{\tabcolsep}{4pt}
\resizebox{\textwidth}{!}{\begin{tabular}{ll | *{9}{cc|}}
\toprule
 & \multirow{2}{*}{\textbf{Eval Method}}
& \multicolumn{2}{c}{\textbf{HE-PY}}
& \multicolumn{2}{c}{\textbf{HE-CPP}}
& \multicolumn{2}{c}{\textbf{HE-Go}}
& \multicolumn{2}{c}{\textbf{HE-JS}}
& \multicolumn{2}{c}{\textbf{HE-JA}}
& \multicolumn{2}{c}{\textbf{APPS}}
& \multicolumn{2}{c}{\textbf{BCB}} 
& \multicolumn{2}{c}{\textbf{DB}}
& \multicolumn{2}{c}{\textbf{VP}}
\\
\cmidrule(lr){3-4} \cmidrule(lr){5-6} \cmidrule(lr){7-8}
\cmidrule(lr){9-10} \cmidrule(lr){11-12} \cmidrule(lr){13-14}
\cmidrule(lr){15-16} \cmidrule(lr){17-18} \cmidrule{19-20}
& 
& F1 & Acc
& F1 & Acc
& F1 & Acc
& F1 & Acc
& F1 & Acc
& F1 & Acc
& F1 & Acc 
& F1 & Acc 
& F1 & Acc 
\\
\midrule
\multirow[c]{9}{*}{\rotatebox[origin=c]{90}{GPT-4o-mini}}
   & Vanilla & 6.0 & \secondbest{66.5} & 18.1 & 54.1 & 23.7 & 55.4 & 25.5 & 55.7 & 30.0 & 57.6 & 5.5 & 50.9 & 12.3 & 49.0 & 15.9 & 54.3 & 1.1 & \best{49.7} \\
   & ICE-Score & 37.4 & 60.0 & 22.2 & 54.9 & 31.7 & 57.4 & 27.2 & 55.9 & 35.0 & 58.8 & 15.5 & 51.2 & 20.8 & 44.4 & 38.8 & 60.3 & 9.7 & 43.2 \\
   & CodeJudge & 23.3 & 55.3 & 11.9 & 52.7 & 17.2 & 52.5 & 12.6 & 51.7 & 8.4 & 49.2 & 18.6 & \best{53.2} & 18.0 & 50.3 & 40.1 & 59.1 & 18.1 & 46.2 \\
   & 2PR & 27.1 & 56.6 & 19.8 & 54.9 & 22.7 & 54.8 & 24.2 & 55.0 & 24.6 & 55.6 & 6.7 & 50.5 & 24.3 & 54.4 & 13.0 & 53.2 & 1.7 & 48.7 \\
   & BC & 43.2 & 61.5 & 26.7 & 55.6 & 33.3 & 57.3 & 35.0 & 58.0 & 36.6 & 59.4 & 13.4 & \secondbest{51.4} & 39.1 & 56.0 & 45.5 & \best{62.2} & 12.4 & 46.8 \\
   & MCTS-Judge & \secondbest{46.4} & 61.2 & 30.0 & 55.2 & 38.4 & 58.4 & 35.2 & 57.2 & 34.5 & 58.3 & 42.8 & 50.6 & 42.5 & 58.0 & -- & -- & -- & -- \\
   & CA-SP & 40.3 & 60.8 & 23.4 & 54.7 & 31.5 & 57.1 & 32.1 & 57.6 & 35.1 & 59.0 & 12.6 & 50.5 & 33.3 & 54.4 & 44.0 & \secondbest{62.0} & 12.0 & 46.8 \\
   & \gcell CoCoA Best $\lambda$ & \gcell \best{72.6} & \gcell \best{71.5} & \gcell \best{66.3} & \gcell \best{67.4} & \gcell \best{68.2} & \gcell \best{66.4} & \gcell \best{67.6} & \gcell \best{66.5} & \gcell \best{72.9} & \gcell \best{70.8} & \gcell \best{66.7} & \gcell 51.2 & \gcell \best{69.6} & \gcell \best{62.1} & \gcell \best{68.6} & \gcell 58.5 & \gcell \best{65.5} & \gcell \secondbest{49.3} \\
   & \gcell CoCoA Worst $\lambda$ & \gcell 43.5 & \gcell 61.0 & \gcell \secondbest{44.9} & \gcell \secondbest{60.6} & \gcell \secondbest{50.6} & \gcell \secondbest{60.8} & \gcell \secondbest{40.9} & \gcell \secondbest{58.5} & \gcell \secondbest{49.3} & \gcell \secondbest{60.6} & \gcell \secondbest{44.0} & \gcell 50.0 & \gcell \secondbest{62.3} & \gcell \secondbest{61.9} & \gcell \secondbest{53.0} & \gcell 59.1 & \gcell \secondbest{34.3} & \gcell 44.3 \\
\midrule
\multirow[c]{9}{*}{\rotatebox[origin=c]{90}{Qwen-2.5-7B}}
   & Vanilla & 42.2 & \secondbest{58.9} & 34.9 & 55.9 & \secondbest{45.8} & \best{58.8} & 37.4 & 56.3 & 39.9 & 56.8 & 3.4 & 50.0 & 32.5 & 47.8 & 25.8 & 55.0 & 4.7 & 46.7 \\
   & ICE-Score & 44.6 & 58.4 & 35.3 & 55.4 & 45.2 & \secondbest{57.7} & 38.9 & \secondbest{57.0} & 37.4 & 56.2 & 13.9 & 49.1 & 12.4 & 37.6 & 27.5 & 55.0 & 16.7 & 44.8 \\
   & CodeJudge & 47.3 & 58.7 & \secondbest{42.5} & 56.6 & \best{48.3} & \secondbest{57.7} & 41.5 & \best{57.2} & 43.2 & \secondbest{57.7} & \secondbest{18.9} & \secondbest{51.1} & 17.0 & 33.7 & 29.4 & 55.7 & \secondbest{18.5} & 39.8 \\
   & 2PR & 42.9 & 57.4 & 35.7 & 56.2 & 37.4 & 55.1 & 31.6 & 54.8 & 36.8 & 56.0 & 6.5 & 48.2 & 35.6 & 47.1 & 28.1 & 55.0 & 13.5 & 44.2 \\
   & BC & 43.9 & 58.4 & 27.7 & 55.1 & 39.9 & 55.4 & 38.7 & 56.5 & 37.5 & 55.7 & 1.4 & 49.6 & 27.9 & 43.9 & 13.2 & 52.4 & 2.2 & \secondbest{48.0} \\
   & CA-SP & \secondbest{47.4} & \secondbest{58.9} & 38.9 & \secondbest{57.1} & \secondbest{45.8} & 55.9 & \secondbest{42.2} & \best{57.2} & \secondbest{45.8} & 57.6 & 14.6 & 47.9 & \secondbest{45.7} & \secondbest{51.2} & \secondbest{45.6} & \best{59.4} & 16.1 & 43.9 \\
   & \gcell CoCoA Best $\lambda$ & \gcell \best{48.4} & \gcell \best{60.5} & \gcell \best{47.3} & \gcell \best{57.8} & \gcell 45.7 & \gcell 56.0 & \gcell \best{44.0} & \gcell 55.7 & \gcell \best{52.9} & \gcell \best{60.2} & \gcell \best{52.8} & \gcell 48.0 & \gcell \best{57.4} & \gcell \best{55.1} & \gcell \best{59.6} & \gcell \secondbest{58.4} & \gcell \best{47.5} & \gcell 44.0 \\
   & \gcell CoCoA Worst $\lambda$ & \gcell 35.2 & \gcell 57.3 & \gcell 24.5 & \gcell 53.2 & \gcell 30.5 & \gcell 53.5 & \gcell 30.7 & \gcell 54.8 & \gcell 29.6 & \gcell 53.7 & \gcell 17.5 & \gcell \best{51.2} & \gcell 35.5 & \gcell \secondbest{51.2} & \gcell 27.7 & \gcell 52.5 & \gcell 17.6 & \gcell \best{48.1} \\
\midrule
\multirow[c]{9}{*}{\rotatebox[origin=c]{90}{Llama-3.1-8B}}
   & Vanilla & 33.9 & 56.6 & 11.2 & 51.9 & 21.7 & 53.2 & 18.5 & 53.9 & 31.9 & 55.1 & 28.6 & 49.1 & 19.9 & 49.5 & 33.8 & \secondbest{56.7} & 39.4 & \secondbest{53.5} \\
   & ICE-Score & 47.8 & \secondbest{58.7} & 33.1 & 55.6 & 42.2 & \secondbest{56.0} & 39.2 & 55.4 & 44.6 & 58.2 & 53.7 & 51.2 & 42.0 & 48.8 & 60.4 & \best{58.2} & 61.2 & 50.3 \\
   & CodeJudge & 24.1 & 54.2 & 3.9 & 50.7 & 11.8 & 51.5 & 15.6 & 52.8 & 16.6 & 52.0 & 26.9 & 47.5 & 16.6 & 50.3 & 31.8 & 53.9 & 35.3 & 49.9 \\
   & 2PR & 12.7 & 51.3 & 3.9 & 50.5 & 9.7 & 51.2 & 5.1 & 51.1 & 21.6 & 51.7 & 11.0 & 50.9 & 4.9 & 47.1 & 20.2 & 52.7 & 32.7 & 51.6 \\
   & BC & 55.2 & \best{60.0} & 46.5 & 54.9 & 54.2 & \best{58.3} & 46.2 & 54.3 & 55.9 & \secondbest{58.6} & \secondbest{61.1} & \best{53.2} & 55.2 & \secondbest{57.8} & 57.2 & 55.4 & 60.9 & \best{54.8} \\
   & CA-SP & 57.8 & \secondbest{58.7} & 53.4 & 54.6 & 51.7 & 54.8 & 53.7 & 55.4 & \secondbest{59.9} & \secondbest{58.6} & 60.1 & \secondbest{52.5} & 57.5 & \best{58.0} & \secondbest{63.0} & 56.3 & \secondbest{63.4} & 52.6 \\
   & \gcell CoCoA Best $\lambda$ & \gcell \best{67.2} & \gcell 58.4 & \gcell \best{62.9} & \gcell \secondbest{56.8} & \gcell \best{64.7} & \gcell 54.5 & \gcell \best{65.4} & \gcell \secondbest{57.8} & \gcell \best{65.2} & \gcell 56.6 & \gcell \best{64.8} & \gcell 50.2 & \gcell \best{66.9} & \gcell 54.1 & \gcell \best{67.6} & \gcell 52.8 & \gcell \best{66.0} & \gcell 50.4 \\
   & \gcell CoCoA Worst $\lambda$ & \gcell \secondbest{59.5} & \gcell 58.2 & \gcell \secondbest{54.9} & \gcell \best{57.6} & \gcell \secondbest{55.4} & \gcell 55.7 & \gcell \secondbest{57.0} & \gcell \best{58.3} & \gcell 59.1 & \gcell \best{59.4} & \gcell \secondbest{61.1} & \gcell 49.5 & \gcell \secondbest{64.8} & \gcell 49.7 & \gcell 62.2 & \gcell 50.0 & \gcell 59.7 & \gcell 47.2 \\
\bottomrule
\end{tabular}}
\caption{\textbf{Main Results, CoCoA is a more accurate and balanced correctness evaluator.} For CoCoA, we report the values of the best and worst $\lambda$ in terms of F1 for each model and dataset combination. We report all $\lambda$ values in Table \ref{tab:lambda_results} in the Appendix. Highest values per metric, dataset, and model setting combination are \best{green}, second best is \secondbest{black}. Due to the amount of sequential calls of MCTS-Judge and computational resources, we only compare with it using GPT-4o-mini on a subset of the datasets.}
\label{tab:eval_results}
\end{table}

\subsection{Main Results (RQ1).}

\begin{figure}[h]
    \centering
    \includegraphics[width=\linewidth]{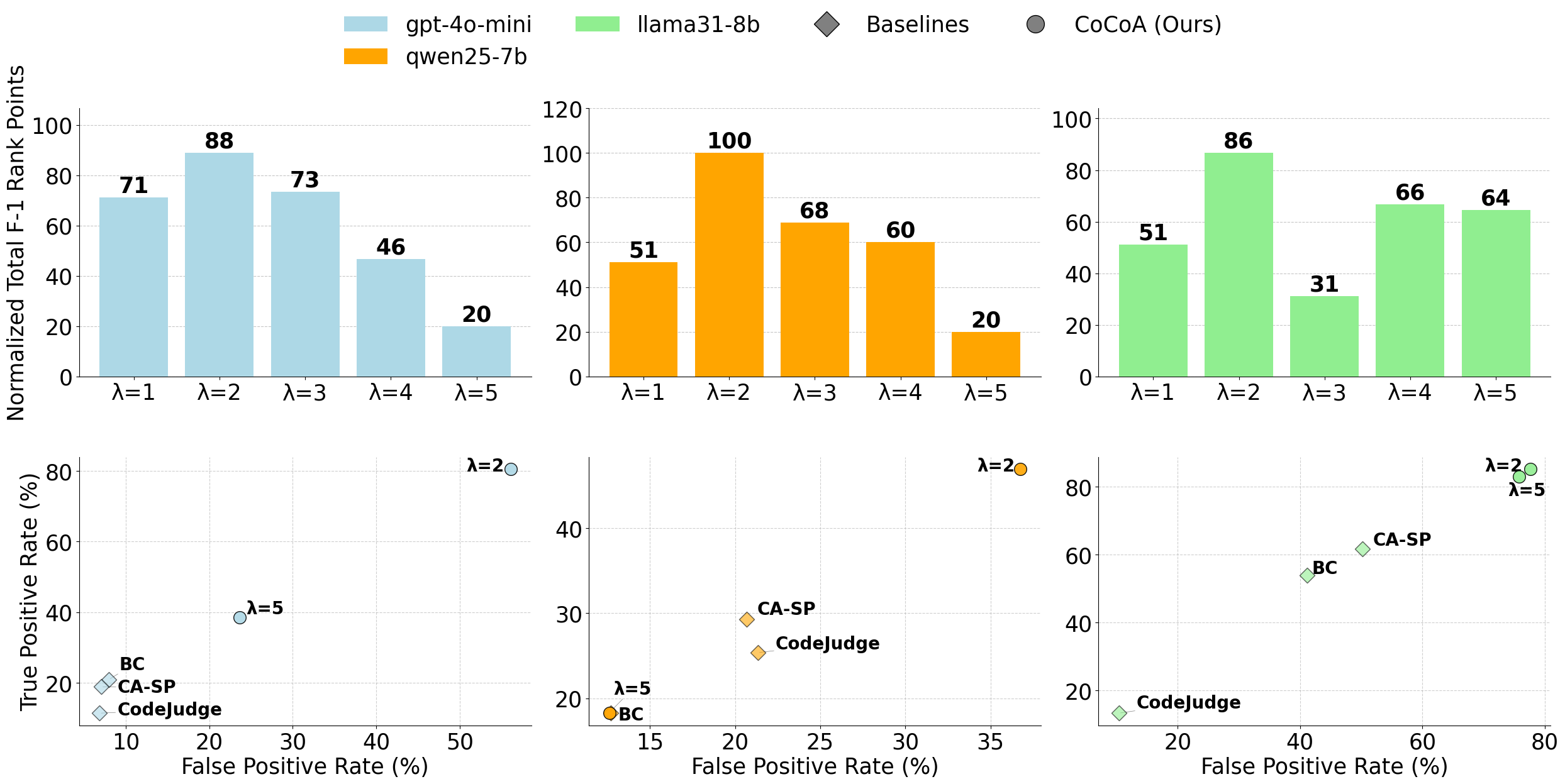}
    \caption{\textbf{Level of Detail Impacts F1.} \textbf{[TOP ROW]} Normalized total F1 ranking points across $\lambda$ values. $\lambda = 2$ across models achieves the most frequently the highest F1, and $\lambda =5$ frequently achieves the lowest. \textbf{[BOTTOM ROW]} Scatter plots per model of True Positive Rate and False Positive Rate for the different evaluation methods.}
    \label{fig:tpr_fpr}
\end{figure}

\paragraph{CoCoA is a more accurate and balanced classifier because it judges on the explanation space, not the implementation space.} In Table \ref{tab:eval_results},  consistently across models and datasets, CoCoA has a significantly higher F1 score with a comparable or higher accuracy to baseline evaluation methods. Figure \ref{fig:tpr_fpr} shows that the True Positive Rates for baselines across models are lower than those for CoCoA, showing that CoCoA evaluations are less biased to false negative judge code as seen in prior work \citep{tong2024codejudge, jin2025uncovering}. From manual inspection, we see many cases like in Figure \ref{fig:example_case} where our intermediate explanation more reliably reflects the functionality of the code being judged. 

\begin{figure}
    \centering
    \includegraphics[width=\linewidth]{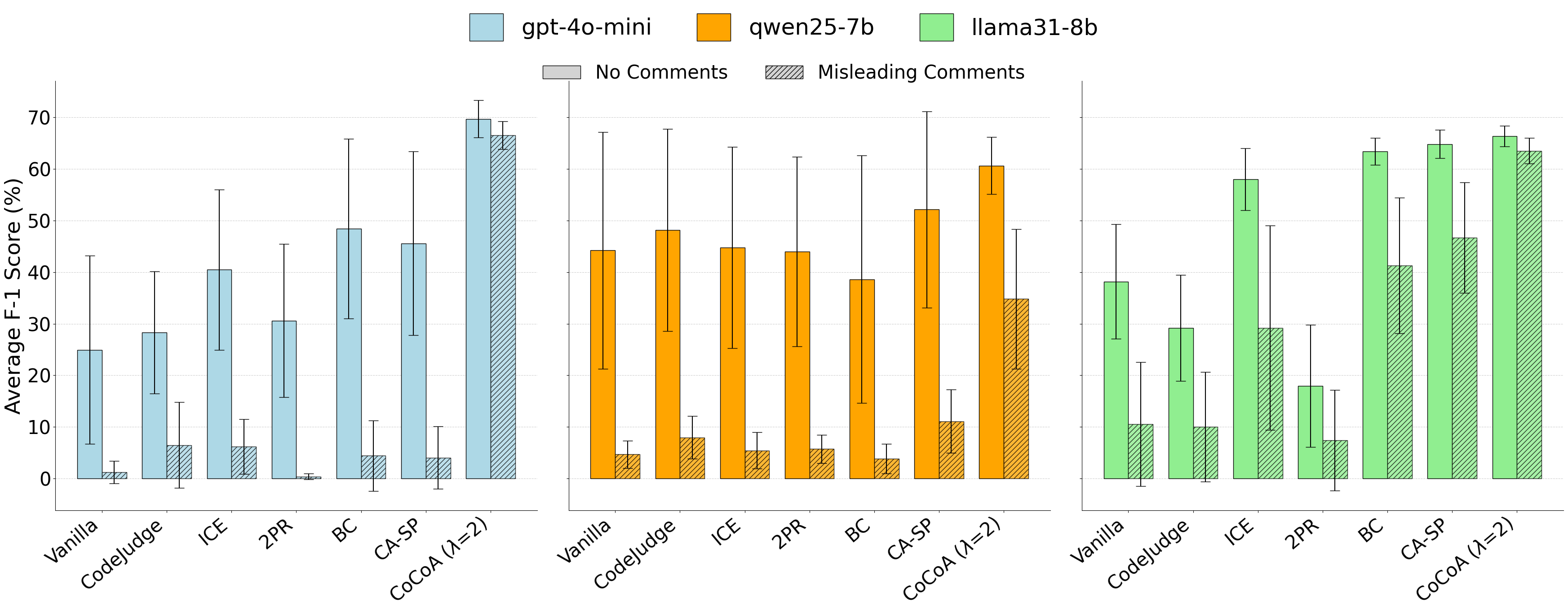}
    \caption{\textbf{CoCoA is robust to Misleading Comments.} Per model across the different evaluation methods, we compare the F1 on the subset of code implementation without comments ($c_t^*$ and $c_t^-$) with the F1 on their counterparts with misleading code comments. We report the average F1 scores across all $9$ datasets.}
    \label{fig:robustness}
\end{figure}

\begin{table}
    \centering
    \resizebox{\textwidth}{!}{\begin{tabular}{c|c|c|c|c|| c|c|c|c}
        Eval Method         & Syntax & Logic & Reference & Multiple & Syntax & Logic & Reference & Multiple \\ \hline
        Eval Model & \multicolumn{4}{c||}{GPT-4o-mini} & \multicolumn{4}{c}{Qwen2.5-7b-Turbo}  \\ \hline
        CodeJudge & 39.7 & 39.4 & 41.4 & 39.6 & 34.6 & 28.1 & 23.5 & 30.3  \\ \hline
        BC & 48.5 & 40.3 & 42.0 & 51.3 & 20.2 & 13.5 & 4.4 & 13.3 \\ \hline
        CA-SP  & 47.0  & 43.4  & 44.4  & 40.7 & \secondbest{45.5} & \secondbest{44.7} & \secondbest{45.0} & \secondbest{47.5} \\ \hline
        \gcell CoCoA Best $\lambda$ & \gcell  \best{68.4} & \gcell \best{68.5} & \gcell  \best{71.0} & \gcell \best{66.0} &  \gcell \best{63.8}  & \gcell \best{56.4}   & \gcell  \best{60.1} & \gcell \best{57.5 }  \\
        \gcell CoCoA Worse $\lambda$ & \gcell  \secondbest{53.0} & \gcell \secondbest{46.8} & \gcell  \secondbest{50.3} & \gcell \secondbest{62.8} &  \gcell 27.9  & \gcell 32.5   & \gcell  29.3 & \gcell 19.6  \\
    \end{tabular}}
    \caption{\textbf{F1 per Error Category on DebugBench.}}
    \label{tab:error_type}
\end{table}

\paragraph{Level of detail in explanation impacts reliability of evaluation.} The top row of Figure \ref{fig:tpr_fpr} shows the normalized cumulative F1 ranking points across $\lambda$ values. Specifically, for each dataset and model setting in Table \ref{tab:eval_results}, we assigned points to the $\lambda$ values based on their F1-score order. The $\lambda$ with the highest F1 score receives $5$ points, the second-highest receives $4$ points, and so on. Figure \ref{fig:tpr_fpr} shows, per model, the total amount of points obtained by a $\lambda$ level normalized by the total amount of points possible, i.e., $45$. We see that across models, $\lambda = 2$ achieves the highest cumulative F1 ranking points. We explicitly prompt for $\lambda = 2$ to not speculate about edge cases to avoid situations seen in \citet{tong2024codejudge, jin2025uncovering}. When $\lambda = 5$, the explanations have the most details and achieve the closest performance to baseline methods. The prompts for $\lambda = 4$ and $\lambda = 5$ are nearly the same, except we specify in $\lambda = 5$ to think of unhandled edge cases in its explanation. We report the F1 and accuracy for all $\lambda$ values for all datasets and models in Table \ref{tab:lambda_results}. The bottom row plots the average True Positive Rate over the average False Positive Rate across datasets per model. We see that generally, the more implementation details given in the explanation lead to closer TPR and FPR to baselines, showing that implementation details lead to higher false negatives. Across models, the effect of different $\lambda$ levels on downstream F1 judgment can vary widely. This result suggests that different models exhibit different levels of permissiveness and conservativeness. We provide example explanations of $\lambda=1,2,5$ in Figure \ref{fig:lambda_examples} in the Appendix.


\paragraph{CoCoA is more robust to misleading comments.} Figure \ref{fig:robustness} shows the difference in the average F1 scores across all datasets between the evaluation methods on code with no comments ($c_t^*$ and $c_t^-$) and the average F1 on code with misleading comments using all three models. CoCoA ($\lambda = 2$) shows stronger robustness to misleading comments than the baseline evaluation methods. This result shows that first explaining the code and evaluating the explanation can remove implementation noise for a more reliable evaluation. CoCoA with high-level explanations is more likely than baseline methods to treat two code implementations the same if they have the same functionality. Per model, we also observe smaller variation with CoCoA ($\lambda =2$) for both no comments and misleading comments compared to baselines, showing the robustness is more consistent across datasets. Examples given in Figure \ref{fig:misleading_task_canonical} in the Appendix.

\paragraph{Code comprehension becomes increasingly more important as code complexity increases.} Similar to the findings of \citet{tong2024codejudge}, we notice that there is a drop in performance across baseline methods and CoCoA when testing on more complex tasks, such as in APPS and VP, which are competition and interview-level questions as opposed to HE. Within HE, we also see that the evaluation methods are generally stronger on HE-PY and HE-JAVA than the other language variants, like in \citet{tong2024codejudge}. 

\paragraph{CoCoA improves performance across various error types.} While most of the datasets (e.g., HE, APPS, and BCB) are perturbed to have $c^-_t$ with logical errors, we also wish to examine a wider variety of errors. Table \ref{tab:error_type} decomposes the F1 achieved on DebugBench into its four categories of errors: Syntax, Logic, Reference, and Multiple. CoCoA yields consistent improvements across all error types. While gains are achieved for syntax errors, they are not disproportionately larger than those on semantic error types, logic and reference. The comparable or greater improvements on semantic error types suggest that CoCoA improves behavior-level representations.

\subsection{Ablations (RQ2).}

Below are findings from ablation experiments that are possible specifically due to separating the comprehension and evaluation into sequential LLM calls, unlike baselines like CA-SP.

\begin{figure}
    \centering
    \includegraphics[width=\linewidth]{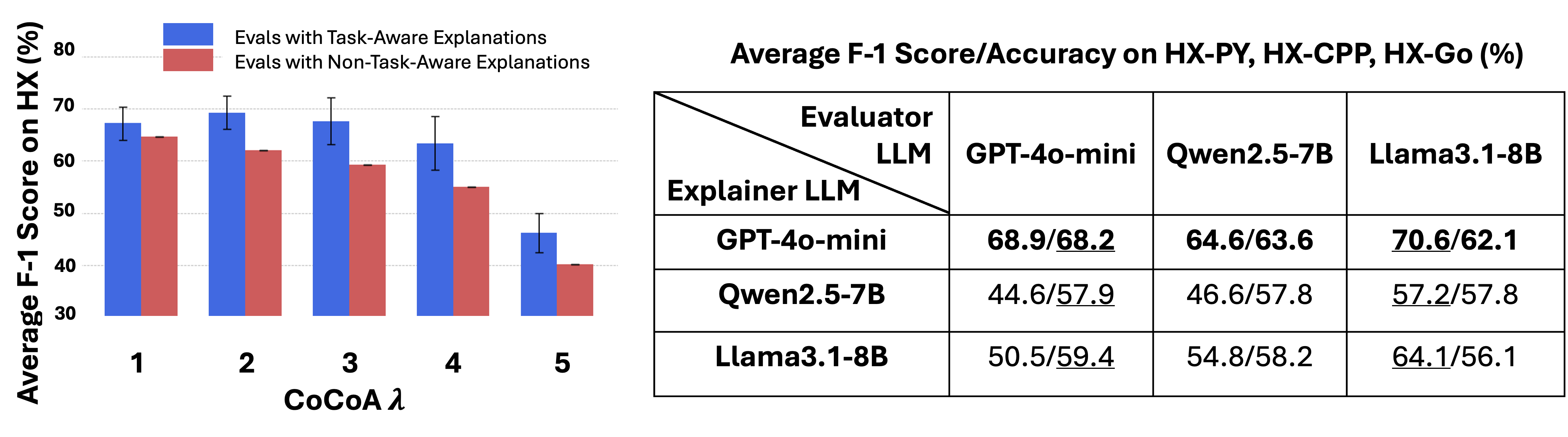}
    \caption{\textbf{Ablations with CoCoA modularity.} \textbf{[LEFT]} Average F1 on all HE datasets across lambda levels comparing between generating explanations with or without the task description. \textbf{[RIGHT]} We use different models for the explainer and evaluator steps. We report the average F1 and accuracy on HE-PY, HE-CPP, and HE-GO. We \textbf{bold} the highest F1 and accuracy per column (evaluator and \underline{underline} the highest per row (explainer).}
    \label{fig:qa_non_qa_model_ablation}
\end{figure}


\paragraph{Task-aware Explanations Help with Judgment.} On the HE datasets, we studied the effect of withholding the task description from the explainer $\pi_e$ so that it only has the implementation to base its explanation. Figure \ref{fig:qa_non_qa_model_ablation} shows that across $\lambda$ values, the F1 scores are up to $\sim 10\%$ higher when the $\pi_e$ has the task. These results suggest that providing context of the code implementation's purpose can help the explainer understand the functionality better, leading to more reliable judgments by $\pi_j$.

\paragraph{Explainer Model Largely Controls the Final Evaluation.} The right of Figure \ref{fig:qa_non_qa_model_ablation} shows the average F1/accuracy across HE-PY, HE-CPP, and HE-Go with CoCoA ($\lambda = 2$) using different LLMs backbones as the explainer $\pi_e$ and evaluator $\pi_j$. With GPT-4o-mini as the explainer, the values are the highest and the row has the lowest variation across explainer models (columns), $\sim6\%$, suggesting the strength of GPT-4o-mini in understanding and explaining code functionality. For Qwen-2.5 and Llama-3.1 explainer models, using a Llama-3.1 evaluator model has a significantly higher F1 score with comparable accuracy than using the other explainer LLMs. However, using the GPT-4o-mini $\pi_e$, the Llama-3.1 $\pi_j$ achieves similar F1 with worse accuracy, and also the lowest F1 score ($63.6 \%$) and accuracy ($62.1\%$) with the GPT-4o-mini $\pi_e$ is still higher than those from other explainer models. These results show that the explainer LLM largely controls the evaluation performance, which makes intuitive sense, and $\pi_e$ is what generates the input to $\pi_j$.


\subsection{Downstream Code Refinement (RQ3).}\label{sec:cr}
We investigate the impact of a balanced evaluator on code refinement. We only activate a refinement step when the verdict is ``Incorrect''. We can better model the real-world issue that overly critical coding agents can be inefficient and even break working code, whereas overly permissive agents may fail to fix existing bugs. We define a code snippet as ``correct after refinement'' as passing all test cases. We report the \textbf{Recovery Rate (RR)}, the number of initially incorrect solutions that became correct after refinement divided by the number of refinement steps, and the \textbf{Preservation Rate (PR)}, the number of initially correct solutions that stayed correct after refinement divided by the number of refinement steps. An RR of $1$ means that every refinement step led to a correct fix. A PR of $1$ means that there was a non-breaking refinement step for each correct solution, and a PR below $1$ means that, on average, refinement broke the working code. We use \textit{GPT-4o-mini} for the refiner agent that takes in the task, code snippet, and evaluation to produce updated code. In Table \ref{tab:refine_results_new} for HE-PY, we see that CoCoA achieves much higher PR and comparable RR, as the refinement step is triggered less than with baseline evaluators.

\begin{table}[!h]
\centering
\small
\setlength{\tabcolsep}{4pt}
\resizebox{0.6\textwidth}{!}{
\begin{tabular}{l | cc  | cc  | cc }
\toprule
 & \multicolumn{2}{c|}{\textbf{GPT-4o-mini}}
& \multicolumn{2}{c|}{\textbf{Qwen-2.5-7B}}
& \multicolumn{2}{c}{\textbf{Llama-3.1-8B}} \\ 

\textbf{Eval Method} & PR & RR  
& PR & RR  
& PR & RR  \\
\midrule


CodeJudge 
& 0.98 & 0.83  
& 2.01 & 0.81 
& 0.83 & 0.85  \\

BC 
& 1.35 & 0.82   
& 1.54 & 0.80   
& 2.19 & 0.75    \\

CA-SP 
& 1.26 & 0.81    
& 1.81 & 0.82   
& 3.51 & 0.81   \\

\gcell CoCoA $\lambda = 2$ 
&\gcell 10.15 & \gcell 0.90  
&\gcell 11.2 & \gcell 0.84   
&\gcell 27.08  & \gcell 0.63    \\

\gcell CoCoA $\lambda = 5$ 
& \gcell 11.27 & \gcell 0.86   
& \gcell 12.5  & \gcell 0.83  
& \gcell 22.91 &  \gcell 0.61   \\

\bottomrule
\end{tabular}
}
\caption{\textbf{Preservation Rate (PR) and Recovery Rate (RR) across models on HE-PY.}}
\label{tab:refine_results_new}
\end{table}

\subsection{Limitation of CoCoA: False Positives vs False Negatives}

Although CoCoA can achieve significant gains in F1 and accuracy, we do notice a consistent increase in FPR when using CoCoA with lower values of $\lambda$, as shown in Figure \ref{fig:tpr_fpr}. We notice that most of these false positives stem from when the explanation is not detailed enough to capture small bugs like perturbed inequality conditions, like $<=$ instead of $<$, and vice versa, the evaluator will then have no way to realize that the bug is there. However, when going to larger $\lambda$ that provides more specific implementation details, the evaluators can fall back to overly-cautious judgments as reported in \citet{tong2024codejudge, jin2025uncovering}, causing an increase in false negatives.


\section{Conclusion}

We introduced a new paradigm for unsupervised LLM code evaluation that separates comprehension from judgment by evaluating in the explanation space rather than over implementations. We then propose CoCoA, which utilizes this idea by first generating an explicit explanation of program behavior and then evaluating alignment conditioned on this explanation. CoCoA removes the need to jointly reason about low-level implementation and high-level correctness. This decoupling leads to more balanced and reliable evaluation, improving F1 while maintaining comparable accuracy. Our results suggest that failures in unsupervised LLM-based code evaluation stem not only from reasoning limitations, but from the choice of representation over which evaluation is performed. We also demonstrate that evaluation in the explanation space improves robustness. Importantly, these gains are not merely due to additional reasoning steps, but from changing the representation used for judgment. Future work may explore extending this paradigm to other domains beyond task-specific code implementations such as repository maintenance.


\section*{Acknowledgments}
We would like to thank Wesley A. Suttle for feedback on ideation.


\bibliography{colm2026_conference}
\bibliographystyle{colm2026_conference}

\appendix

\section{LLM Usage}

We used ChatGPT to help obtain feedback on the draft of this manuscript in terms of writing, specifically providing clearer and more concise wording. We also used LLM to indirectly generate tables by generating code that reads in results and prints out LaTeX code of a formatted table as a string. 

\section{Experimental Details}\label{sec:app_exp_details}

\paragraph{Datasets.} We provide a brief description of each dataset used. The number of tasks sampled per dataset can be found in Table \ref{tab:dataset_details}.\\

HumanEval-X \citep{zheng2023codegeex} is a collection of coding tasks that has implementation solutions in different languages. Like \citet{tong2024codejudge, wang2025mctsjudgetesttimescalingllmasajudge}, we use Python, C++, Go, Java, and JS. It is the multilingual version of HumanEval \cite{chen2021evaluating}.  \\

APPS \citep{hendrycksapps2021} is a Python code generation benchmark, with three levels of difficulty: Beginner, Interview, and Competition. In this work, we sample tasks from the competition level. \\

BigCodeBench \citep{zhuo2024ice} is a Python code generation benchmark that focuses on multiple function calls. \\

DebugBench \citep{tian-etal-2024-debugbench} is a debugging benchmark, created from questions from LeetCode for multiple languages (Python, C++, and Java). For each task, they contain a solution and an incorrect implementation. DebugBench contains instances of various types of errors (syntax, logic, reference, and multiple). We specifically use the DebugBench Hard level.

VibePass \citep{bansal2026vibepass} is a recent unit test generation benchmark containing tasks from LiveCodeBench \citep{jain2024livecodebench} with 1) ``silver solutions'' that pass all test cases but are not guaranteed to be absolutely correct and 2) near-correct, but buggy solutions.

\begin{table}[h]
    \centering
    \begin{tabular}{c|c|c}
        Dataset & No. of Tasks & No. of (Code, Task) Pairs \\ \midrule
        HE-PY & 155 & 629 \\
        HE-CPP  & 148 & 592 \\
        HE-GO & 162 & 648 \\
        HE-JS &  115 & 460\\
        HE-JAVA & 162  & 648 \\
        APPS & 140 &  560\\
       BCB  &  147  & 588\\
       DB  & 179 & 716\\
        VP & 172 & 688 \\
    \end{tabular}
    \caption{Dataset Statistics. No. of (Code, Task) Pairs is No. of Tasks multiplied by $4$. }
    \label{tab:dataset_details}
\end{table}

\paragraph{Baselines.} All baseline methods evaluate correctness with implementation $c$ as a direct input. We compare against a \textbf{Vanilla} evaluation call that takes in the entire snippet $c$ and is prompted to give a binary assessment on correctness with reasoning. \textbf{ICE-Score} \citep{zhuo2024ice} gives a rating from $0$ to $4$, where $4$ indicates totally correct. \textbf{CodeJudge} \citep{tong2024codejudge} first generates a correctness analysis of the code and then generates a binary final verdict based on the analysis. \textbf{MCTS-Judge} performs reasoning rollouts\footnote{due to the amount of sequential calls, we only perform 1 rollout per task and code evaluation} and evaluates with simulated test cases. We view this as a different form of code comprehension as the MCTS-Judge must understand the code to perform its test case simulation. This step is similar to how our CoCoA explainer provides example cases in its explanation. Both \textbf{Two-Phase Reflection (2PR)} and \textbf{Behavior Comparison (BC)} from \citet{jin2025uncovering} are single-call methods that first require the LLM evaluator to explain the task requirements, and then to audit code against the understood requirements. CodeJudge and BC are the most similar prior works to our CoCoA, as they also prompt the LLM to explain the code and compare the explanation against the rubric. Thus, CodeJudge and BC ask the LLM in a single prompt to 1) understand requirements, then 2) understand functionality then 3) evaluate correctness (for CodeJudge, these steps are specified in the analysis prompt). To better compare with CoCoA, which emphasizes code understanding and evaluation, we also provide another baseline \textbf{Comprehension then Auditing-Single Prompt (CA-SP)}, that modifies BC prompt to only comprehend code, then evaluate. This modified baseline will better highlight the improvements gained from separating comprehension and evaluation into separate calls.

\section{Additional Results and Examples} \label{sec:app_results}

\begin{table}[!h]
\centering
\small
\setlength{\tabcolsep}{4pt}
\resizebox{\textwidth}{!}{\begin{tabular}{ll | *{9}{cc|}}
\toprule
 & \multirow{2}{*}{\textbf{$\lambda$ level}}
& \multicolumn{2}{c}{\textbf{HE-PY}}
& \multicolumn{2}{c}{\textbf{HE-CPP}}
& \multicolumn{2}{c}{\textbf{HE-Go}}
& \multicolumn{2}{c}{\textbf{HE-JS}}
& \multicolumn{2}{c}{\textbf{HE-JA}}
& \multicolumn{2}{c}{\textbf{APPS}}
& \multicolumn{2}{c}{\textbf{BCB}} 
& \multicolumn{2}{c}{\textbf{DB}}
& \multicolumn{2}{c}{\textbf{VP}}
\\
\cmidrule(lr){3-4} \cmidrule(lr){5-6} \cmidrule(lr){7-8}
\cmidrule(lr){9-10} \cmidrule(lr){11-12} \cmidrule(lr){13-14}
\cmidrule(lr){15-16} \cmidrule(lr){17-18} \cmidrule{19-20}
& 
& F1 & Acc
& F1 & Acc
& F1 & Acc
& F1 & Acc
& F1 & Acc
& F1 & Acc
& F1 & Acc 
& F1 & Acc 
& F1 & Acc 
\\
\midrule
\multirow[c]{5}{*}{\rotatebox[origin=c]{90}{GPT-4o-mini}}
   &  $\lambda=1$ &  71.0 &  68.2 &  63.8 &  59.6 &  66.4 &  61.4 &  66.8 &  62.6 &  70.2 &  63.4 &  66.7 &  51.2 &  67.3 &  56.0 &  66.0 &  52.0 &  65.5 &  49.3 \\
   &  $\lambda=2$ &  72.6 &  71.5 &  66.3 &  67.4 &  68.2 &  66.4 &  67.6 &  66.5 &  72.9 &  70.8 &  65.2 &  51.2 &  69.6 &  62.1 &  66.9 &  54.5 &  64.9 &  49.4 \\
   &  $\lambda=3$ &  72.0 &  72.1 &  63.6 &  65.0 &  67.5 &  67.0 &  63.1 &  63.9 &  72.2 &  69.9 &  66.1 &  53.9 &  67.1 &  60.9 &  68.6 &  58.5 &  64.7 &  49.6 \\
   &  $\lambda=4$ &  66.0 &  68.5 &  58.1 &  62.7 &  67.2 &  67.4 &  56.9 &  61.5 &  68.3 &  69.3 &  65.8 &  55.4 &  66.7 &  60.5 &  66.8 &  58.0 &  63.9 &  50.6 \\
   &  $\lambda=5$ &  43.5 &  61.0 &  44.9 &  60.6 &  50.6 &  60.8 &  40.9 &  58.5 &  49.3 &  60.6 &  44.0 &  50.0 &  62.3 &  61.9 &  53.0 &  59.1 &  34.3 &  44.3 \\
\midrule
\multirow[c]{5}{*}{\rotatebox[origin=c]{90}{Qwen-2.5-7B}}
   &  $\lambda=1$ &  36.3 &  56.9 &  38.5 &  56.8 &  38.0 &  55.2 &  31.8 &  55.2 &  36.4 &  54.6 &  45.8 &  47.1 &  47.2 &  52.4 &  57.3 &  58.1 &  40.8 &  43.9 \\
   &  $\lambda=2$ &  48.4 &  60.5 &  47.3 &  57.8 &  45.7 &  56.0 &  44.0 &  55.7 &  52.9 &  60.2 &  52.8 &  48.0 &  57.4 &  55.1 &  59.6 &  58.4 &  47.5 &  44.0 \\
   &  $\lambda=3$ &  42.2 &  58.4 &  38.4 &  56.1 &  39.2 &  56.5 &  38.4 &  56.1 &  39.5 &  56.0 &  49.3 &  51.4 &  53.5 &  53.9 &  55.8 &  58.2 &  41.0 &  42.3 \\
   &  $\lambda=4$ &  41.0 &  56.3 &  36.5 &  54.2 &  40.8 &  54.3 &  40.9 &  56.7 &  44.6 &  56.6 &  46.4 &  50.5 &  53.1 &  54.3 &  50.2 &  53.8 &  37.6 &  44.9 \\
   &  $\lambda=5$ &  35.2 &  57.3 &  24.5 &  53.2 &  30.5 &  53.5 &  30.7 &  54.8 &  29.6 &  53.7 &  17.5 &  51.2 &  35.5 &  51.2 &  27.7 &  52.5 &  17.6 &  48.1 \\
\midrule
\multirow[c]{5}{*}{\rotatebox[origin=c]{90}{Llama-3.1-8B}}
   &  $\lambda=1$ &  67.2 &  58.4 &  61.7 &  53.0 &  63.0 &  53.4 &  62.8 &  53.9 &  64.2 &  55.2 &  61.1 &  49.5 &  65.2 &  49.7 &  62.2 &  50.0 &  60.3 &  46.5 \\
   &  $\lambda=2$ &  65.0 &  56.9 &  62.7 &  56.9 &  64.7 &  54.5 &  62.0 &  53.7 &  65.2 &  56.6 &  64.8 &  50.2 &  66.9 &  54.1 &  65.6 &  50.8 &  66.0 &  50.4 \\
   &  $\lambda=3$ &  59.5 &  58.2 &  54.9 &  57.6 &  55.4 &  55.7 &  57.0 &  58.3 &  59.1 &  59.4 &  61.2 &  50.7 &  66.1 &  55.4 &  63.0 &  52.7 &  59.7 &  47.2 \\
   &  $\lambda=4$ &  64.6 &  56.0 &  61.0 &  53.2 &  61.7 &  50.5 &  65.4 &  57.8 &  64.4 &  55.6 &  63.8 &  49.3 &  64.8 &  49.7 &  67.6 &  52.8 &  65.7 &  50.4 \\
   &  $\lambda=5$ &  63.6 &  55.3 &  62.9 &  56.8 &  63.4 &  53.9 &  63.9 &  58.3 &  61.9 &  54.2 &  63.5 &  50.9 &  66.1 &  51.9 &  65.6 &  50.8 &  65.5 &  50.4 \\
\bottomrule
\end{tabular}}
\caption{Results on all datasets and models for each $\lambda$ level.}
\label{tab:lambda_results}
\end{table}

\begin{figure}[!h]
    \centering
    \includegraphics[width=\linewidth]{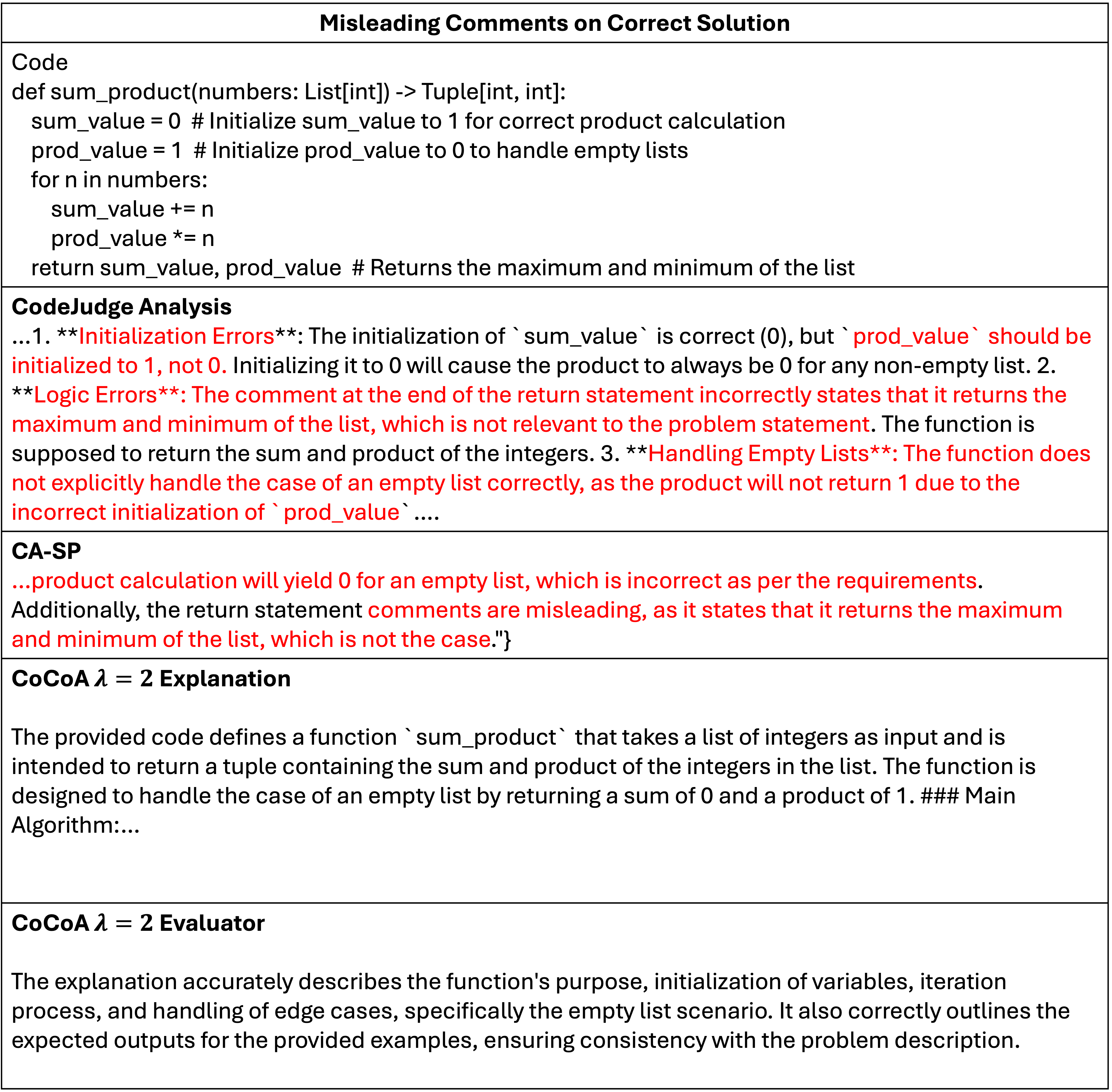}
    \caption{Same solution example as in Table \ref{fig:example_case} but with misleading task comments. We see that CoCoA $\lambda = 2$ is more robust to these comments as they are not taken into account in the explanation. We even see that CA-SP identifies one of the comments as wrong, leading it to an ``Incorrect'' verdict.}
    \label{fig:misleading_task_canonical}
\end{figure}

\begin{figure}[!h]
    \centering
    \includegraphics[width=\linewidth]{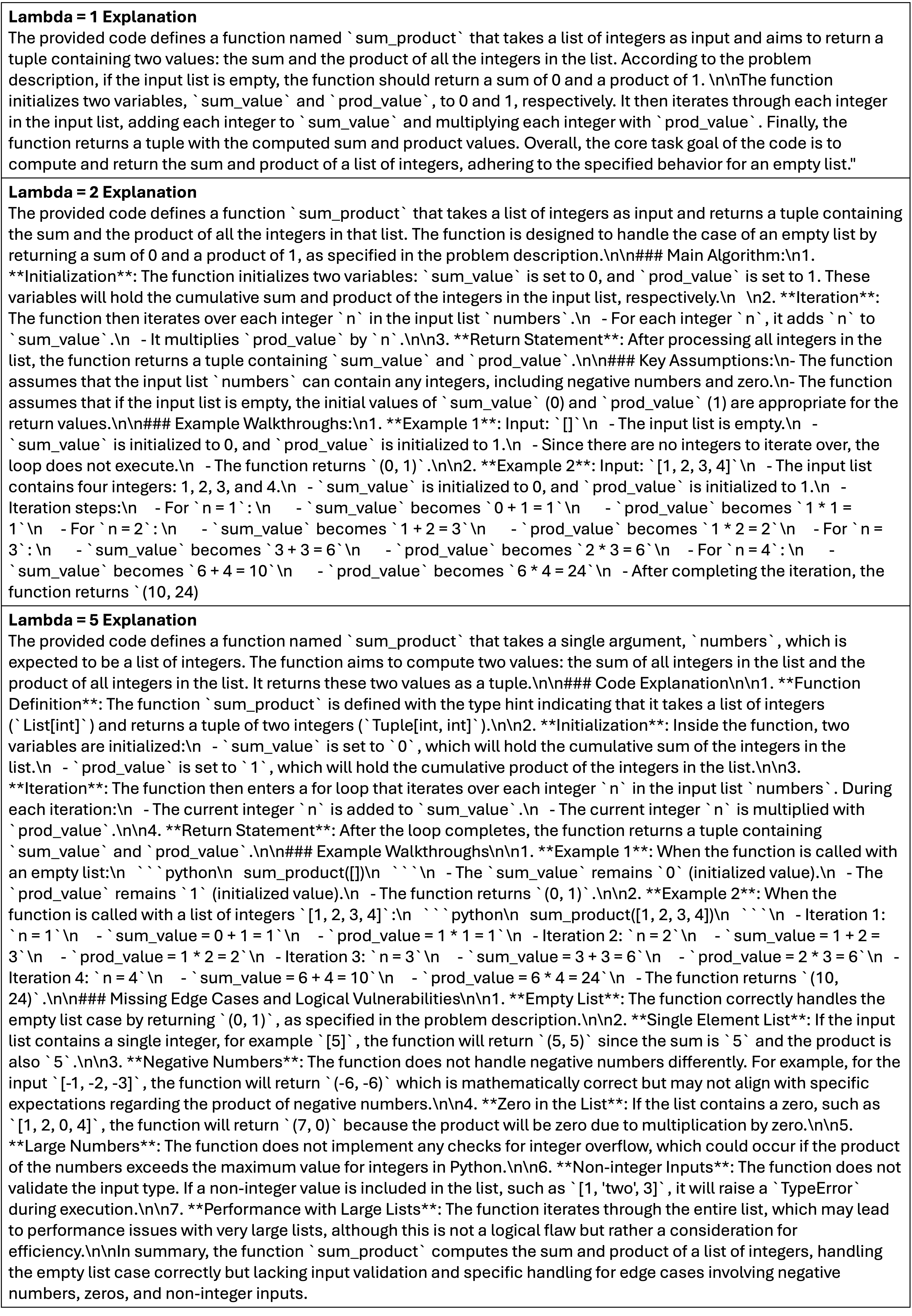}
    \caption{Example CoCoA explanation with different levels of lambda for the code snippet and task in Figure \ref{fig:example_case}. We can see that $\lambda=5$ speculates edge cases (as prompted) like validating input type.}
    \label{fig:lambda_examples}
\end{figure}




\subsection{Prompt Set}\label{sec:prompts}

\begin{figure}[!h]
\centering

\begin{subfigure}{\textwidth}
    \centering
    \includegraphics[width=0.7\textwidth]{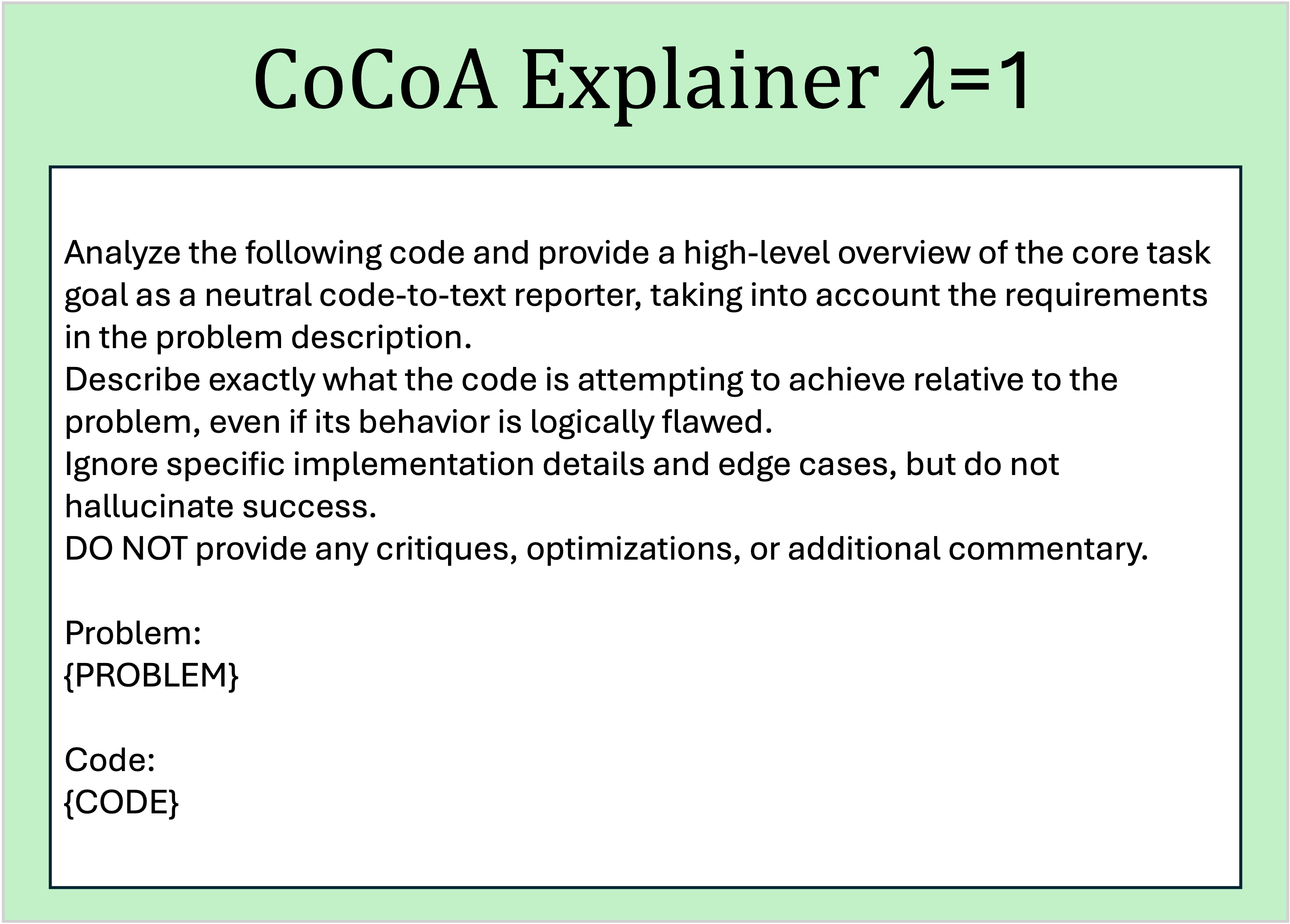}
\end{subfigure}

\begin{subfigure}{\textwidth}
    \centering
    \includegraphics[width=0.7\textwidth]{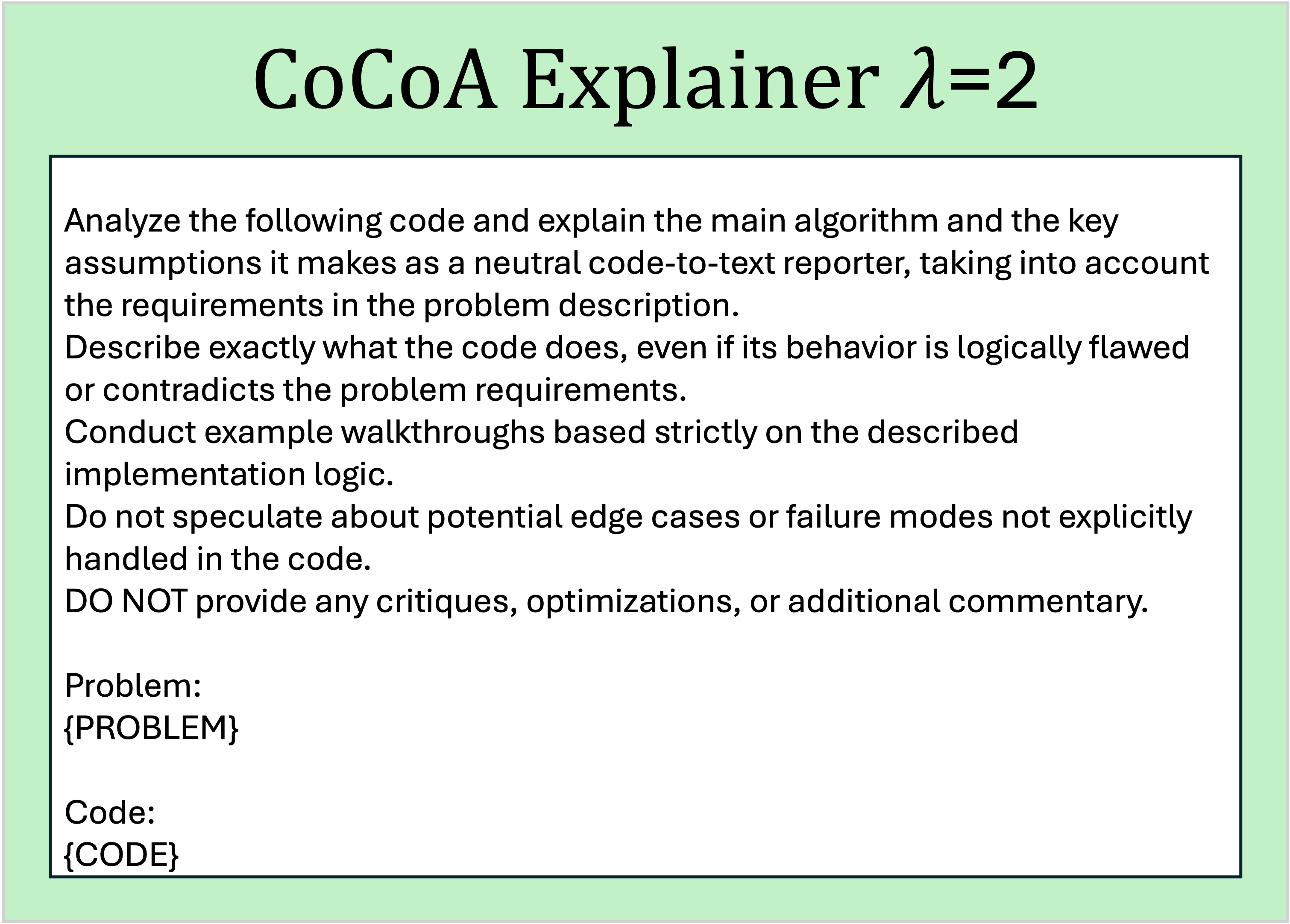}

\end{subfigure}

\label{fig:lam12}
\end{figure}

\begin{figure}[!h]
\centering

\begin{subfigure}{\textwidth}
    \centering
    \includegraphics[width=0.7\textwidth]{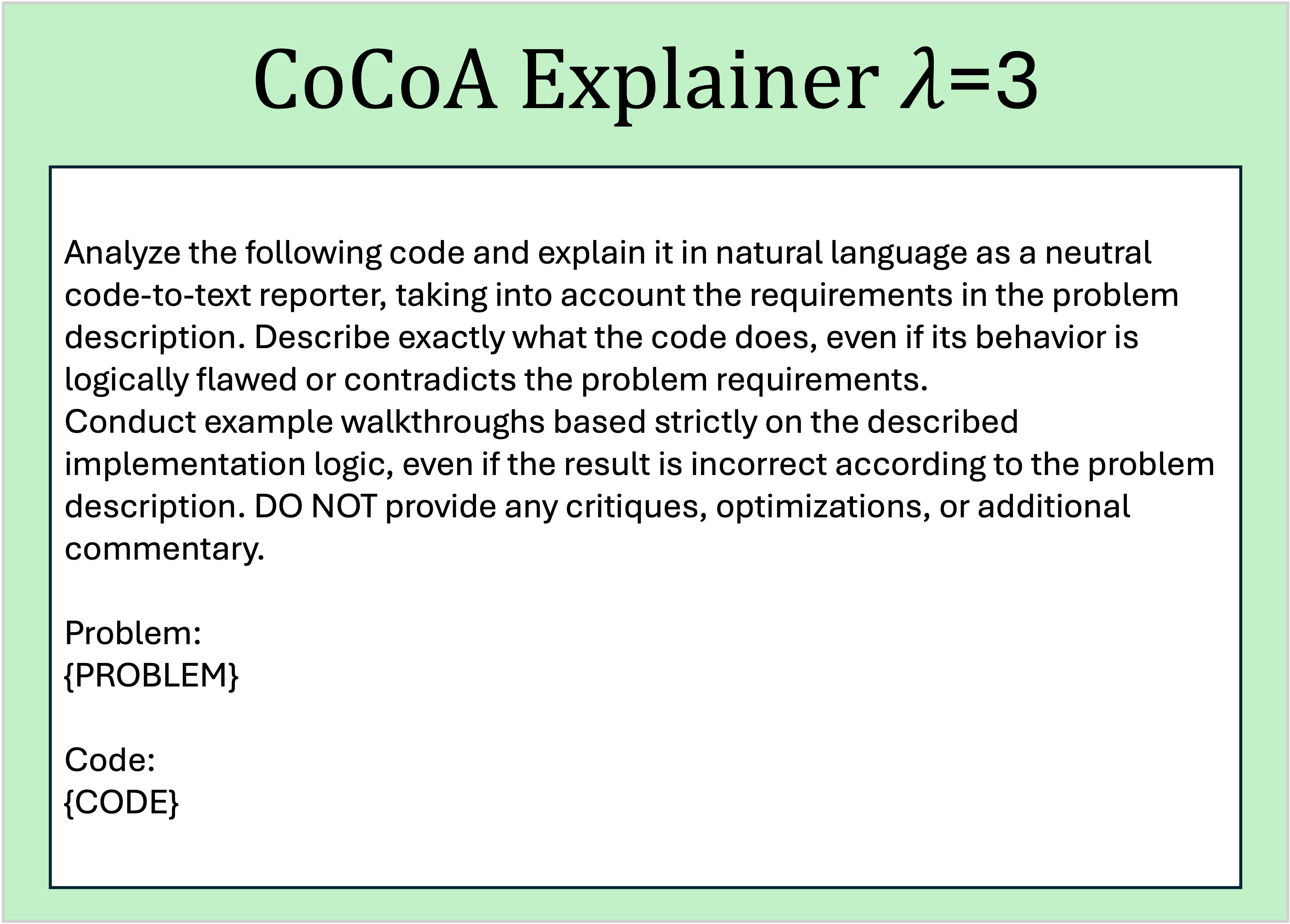}
\end{subfigure}

\begin{subfigure}{\textwidth}
    \centering
    \includegraphics[width=0.7\textwidth]{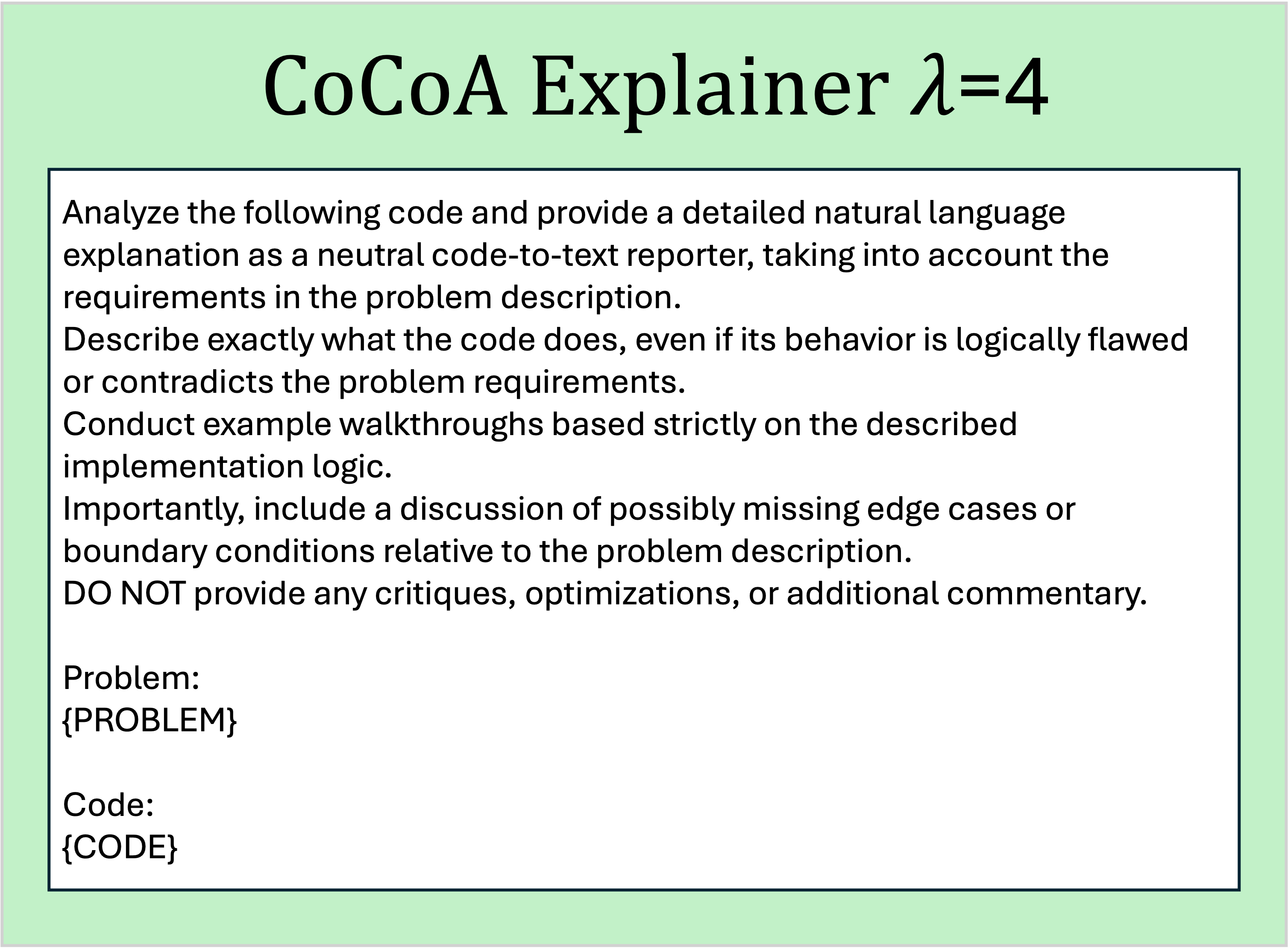}
    
\end{subfigure}

\label{fig:lam34}
\end{figure}

\begin{figure}[!h]
\centering

\begin{subfigure}{\textwidth}
    \centering
    \includegraphics[width=0.7\textwidth]{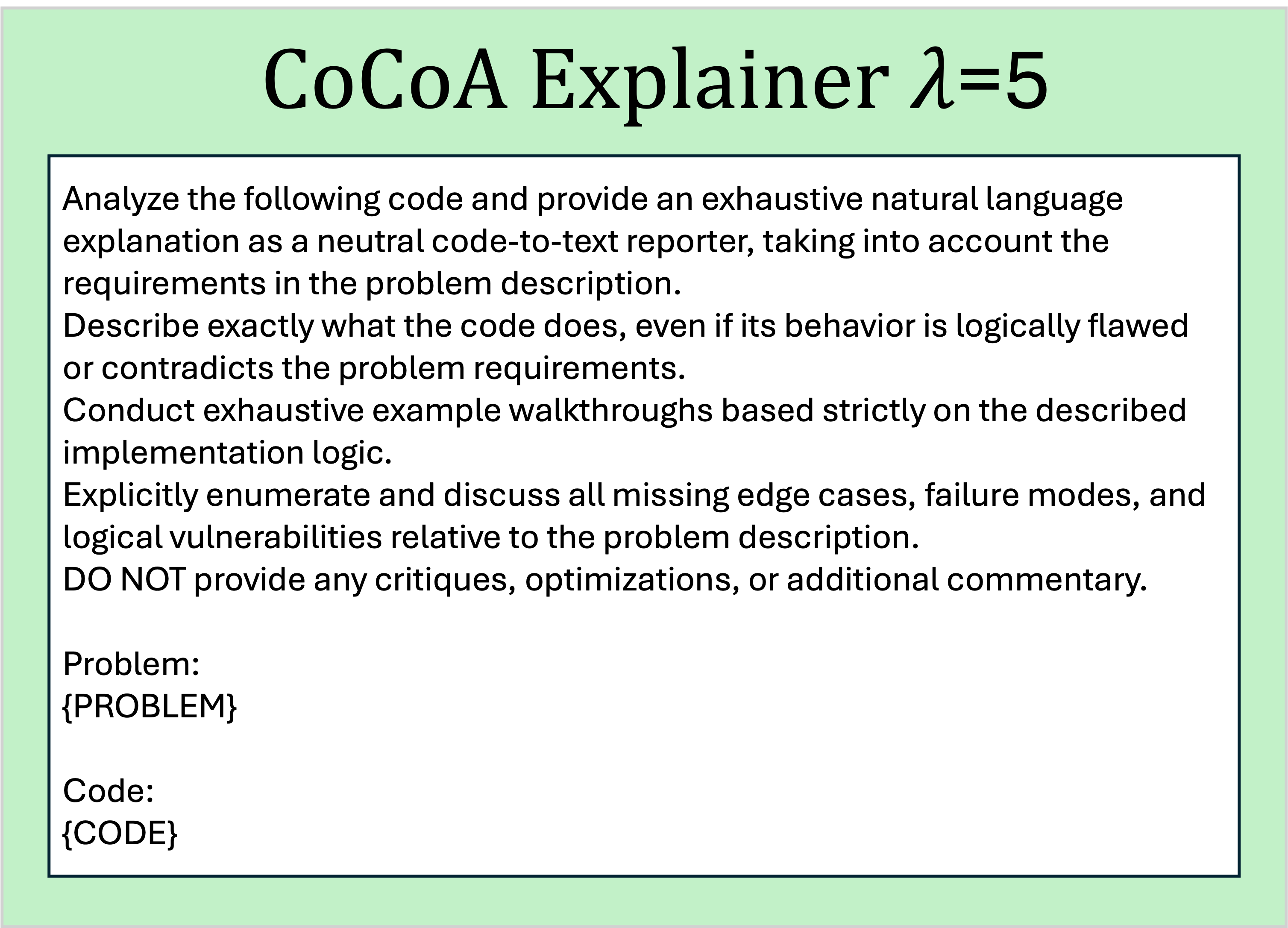}
\end{subfigure}

\begin{subfigure}{\textwidth}
    \centering
    \includegraphics[width=0.7\textwidth]{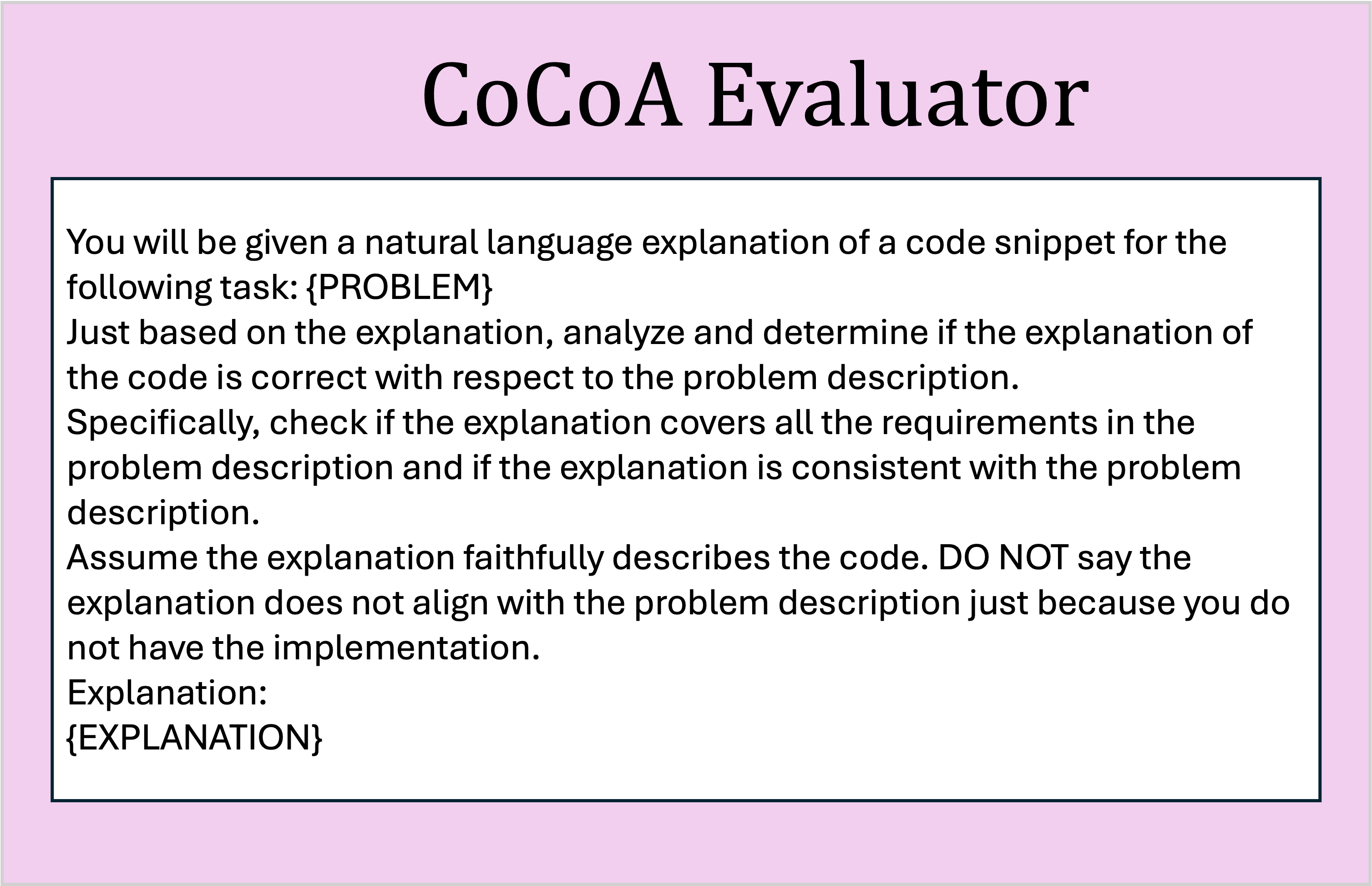}
\end{subfigure}

\label{fig:lam5eval}
\end{figure}

\begin{figure}[!h]
\centering

\begin{subfigure}{\textwidth}
    \centering
    \includegraphics[width=0.7\textwidth]{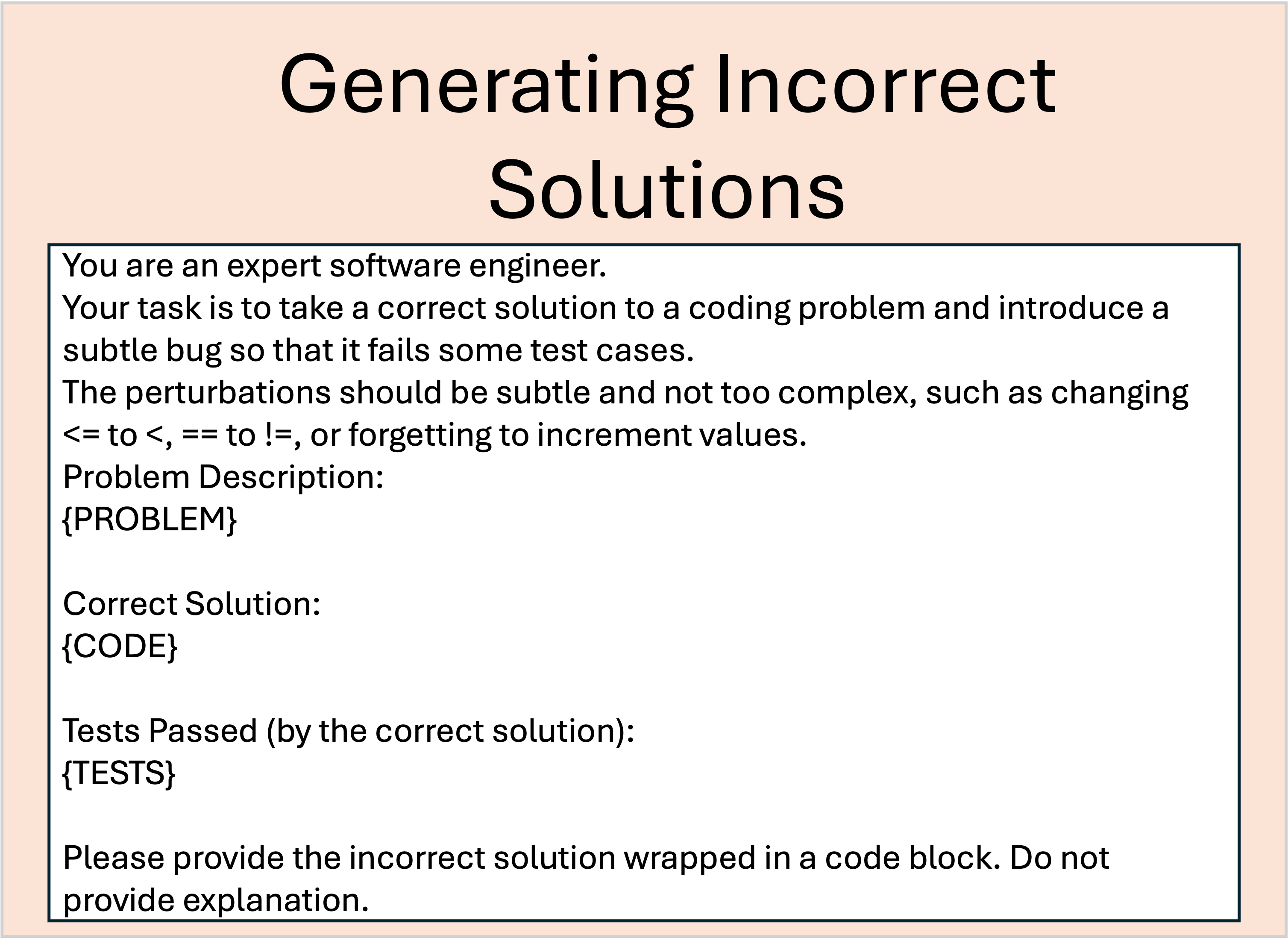}
\end{subfigure}

\label{fig:incorrectsol}
\end{figure}

\end{document}

%% file: math_commands.tex
\usepackage{amsmath,amsfonts,bm}

\def\eqref#1{equation~\ref{#1}}

\def\1{\bm{1}}

\DeclareMathAlphabet{\mathsfit}{\encodingdefault}{\sfdefault}{m}{sl}
\SetMathAlphabet{\mathsfit}{bold}{\encodingdefault}{\sfdefault}{bx}{n}

